# Fuzzy Integer Linear Programming Mathematical Models for Examination Timetable Problem

Arindam Chaudhari[1] and Kajal De[2]

[1]Assistant Professor (Computer Science Engineering),

NIIT University, Neemrana

Rajasthan, India

arindam_chau@yahoo.co.in

[2]Professor in Mathematics,

School of Science,

Netaji Subhas Open University,

Kolkata, India

ABSTRACT. *Examination Timetable Problem (ETP) is NP–Hard combinatorial optimization problem. It has received tremendous research attention during the past few years given its wide use in universities. ETP can be defined as assignment of courses to be examined, candidates to time periods and examination rooms while satisfying a set of constraints which may be either hard or soft. Several methods have been proposed most of which are based on heuristics like Search techniques, Evolutionary Computation etc. In this Paper, we develop three mathematical models for Netaji Subhas Open University, Kolkata, India using Fuzzy Integer Linear Programming (FILP) technique. In most real life situations, information available in is not exact, lacks precision and has an inherent degree of vagueness. To deal with this we model various allocation variables through fuzzy numbers expressing lack of precision the decision maker has. The solution to the problem is obtained using Fuzzy number ranking method. Each feasible solution has fuzzy number obtained by Fuzzy objective function. The different FILP technique performance are demonstrated by experimental data generated through extensive simulation from Netaji Subhas Open University, Kolkata, India in terms of its execution times. The proposed FILP models are compared with commonly used heuristic viz. Integer Linear Programming approach on experimental data which gives an idea about quality of heuristic. The techniques are also compared with different Artificial Intelligence based heuristics for ETP with respect to best and mean cost as well as execution time measures on Carter benchmark datasets to illustrate its effectiveness. FILP paradigm takes an appreciable amount of time to generate satisfactory solution in comparison to other heuristics. The formulation thus serves as good benchmark for other heuristics. The experimental study presented here focuses on producing a methodology that generalizes well over spectrum of techniques that generates significant results for one or more datasets. The performance of FILP model is finally compared to the best results cited in literature for Carter benchmarks to assess its potential. The problem can be further reduced by formulating with lesser number of allocation variables it without affecting optimality of solution obtained. FLIP model for ETP can also be adapted to solve other ETP as well as combinatorial optimization problems. To the best of our knowledge this is first work on ETP using FILP technique.*

1. **Introduction.** In today's fast moving competitive world one of the most fundamental principles is search for an optimization state. Optimization state phenomena are evident in all forms of industrial systems, where achievement of best possible solution is the ultimate resort. One such commonly studied optimization problem is Examination Timetable Problem (ETP) [1–12] which is an important problem in Operations Research and Artificial Intelligence (AI). ETP is an important category of university timetable problem and is one of the most difficult problems faced by universities and colleges today. It is often classified in the family of combinatorial optimization problems [13–16].

ETP can be defined as assignment of courses to be examined, candidates to time periods and examination rooms satisfying a set of constraints while optimizing utilization of existing facilities such that desired objectives are satisfied. The problem constraints may be either hard or soft [17–24]. Hard constraints such as avoiding student examination collision and room over-sizing must be satisfied. Soft constraints such as scheduling large candidate examinations [25–27] first may be tolerable but must be minimized as much as possible. ETP requires allocation of several examination slots of different categories such as theory and practical examinations. Given the increasing number of students in universities, a large number of courses are offered every term. Each course has different number of enrolled students and each classroom has different capacities which make assignment operation complicated. Furthermore, it is not enough to schedule examination of course in classroom with higher capacity than the number of enrolled students, since this can lead to inefficient utilization of classrooms which can cause difficulties for university authorities. The automation of timetable problem is thus an important task as it saves lot of man-hours work to institutions and provides optimal solutions with constraint satisfaction that can boost productivity, quality of examination and services [18, 28]. However, large-scale examination timetables may need many hours of work spent by qualified person or team in order to produce high quality timetables with optimal constraint satisfaction [3, 19–21, 29] and optimization of timetable's objectives.

The real life timetable problems have many forms like education timetable (course and exam), employee timetable, sports timetable, transportation timetable etc. Timetable problems as well as scheduling problems are generally NP–Hard constrained optimization problems [2, 5, 25–26, 30–34], of combinatorial nature and no optimal algorithm is known which generates solution within reasonable time. These problems are mainly classified as constraint satisfaction problems [3, 19–21, 29]. There are number of versions of ETP differing from one university to another [1, 8, 11, 27, 34–53]. ETP differs considerably from the university course scheduling problem, where an examination occupies only one slot throughout the planning period, while course timetable may require several slots and with different categories such as lectures, tutorials and practical sessions. An examination may span several weeks while course timetable must fit within a week, which repeats for the whole semester; an examination room can occupy more than one examination while course schedules cannot.

Over decades scheduling community has used domain knowledge in order to generate high quality solution [1, 3–12, 18–19, 23–28, 33–53]. Therefore, it is very likely that these algorithms could not run on different scheduling problems. However, in the recent past there has been an increase in research towards generating algorithms that work well over a range of scheduling problems [2, 4–5, 8, 11, 13–18, 21–22, 28, 30–32, 37, 42, 50, 54–59], by choosing appropriate heuristics. Such algorithms are usually referred to as hyper-heuristics [6–7, 14–15, 17, 21, 23–24, 35–37, 40–41, 44, 46, 50, 54, 57, 60–61]. In order to consider different scheduling problems, there exists a unique formalism for their representation [2, 8, 13–16, 18, 21–22, 30–32, 54, 56, 62] that will capture variety of domain-specific constraints. In scheduling community there have been attempts to develop language for constraint representation [62–63], but they addressed only a single type of problem. Le Pape [29] presented software schedule which implements constraint language that is powerful enough to represent variety of resource and temporal constraints. However, such representation is more appropriate for certain algorithms. Some other algorithms such as meta-heuristics [24] may need to map given representation of problem into a form suitable for their execution before starting to construct the schedule.

A lot of work has been done on ETP with respect to their studies on specific universities and many formulations and algorithms have been developed. One of the important computing paradigms is graph coloring concept [34, 56, 64] where vertices represent courses and an arc joins two vertices only if they cannot be scheduled at same time. The problem is thus to find chromatic number of resulting graph [56, 64]. However, chromatic number problem is also NP–Hard. Due to complexity of the problem, most of work done concentrates on heuristic algorithms which try to find good approximate solutions [6–7, 14–15, 17, 23–24, 35–37, 40–41, 44, 46, 50, 52, 54, 57, 61]. Some of these include Evolutionary Algorithms, Genetic Algorithms (GA), Ant Colony Optimization, Tabu Search, Simulated Annealing [8, 10, 12, 17, 22, 24, 28, 30, 32–34, 38, 40, 42–43, 45, 47, 50, 54, 58–61], Fuzzy Heuristics [41, 44, 51], Scatter Search [35, 39] methods and host of Heuristic techniques. Heuristic optimization methods are explicitly aimed at good feasible solutions that may not be optimal where complexity of problem or limited time available does not allow exact solution. Generally, two questions arise viz., how fast the solution is computed and how close the solution is to optimal one. Tradeoff is often required between time and quality which is taken care of by running simpler algorithm more than once, comparing results obtained with more complicated ones and effectiveness in comparing different heuristics. The empirical evaluation of heuristic method is based on analytical difficulty involved in problem and pathological worst case result.

In recent past these heuristic tools have been combined among themselves with knowledge elements [33, 37–38, 42, 45, 50, 58–59, 61, 65–66] as well as with more traditional approaches such as Statistical and Fuzzy analysis [67–70] to solve extremely challenging problems. Developing solutions with these tools offers two major advantages viz. shorter development time than traditional approaches and robust systems being insensitive to noisy and missing data. Keeping in view of the recent past, this work attempts to develop three Fuzzy Integer Linear Programming (FILP) mathematical models [68] for ETP at Netaji Subhas Open University, Kolkata, India for which no solution is available presently. As in most real life situations the information available in system is not exact and lack precision and has an inherent degree of vagueness, we consider the various allocation variables as Fuzzy numbers [67, 69] expressing lack of precision that the decision maker has. Fuzzy Logic is a computational paradigm that generalizes classical two-valued logic for reasoning under uncertainty. In order to achieve this, notation of membership in a set needs to become a matter of degree. This is the essence of Fuzzy Sets [67–69]. By doing this, two things are accomplished viz., ease of describing human knowledge involving vague concepts and enhanced ability to develop cost-effective solution to real-world problem. It is multi-valued logic which is model-less approach and clever disguise of Probability Theory. In FILP heuristic technique, each feasible solution has Fuzzy number obtained by Fuzzy objective function [67–69]. The solution to this problem can be obtained using either Fuzzy number ranking method or representation theorem of Fuzzy Sets [67–69]. However, here we restrict ourselves to Fuzzy number ranking method. The performance of different FILP techniques are demonstrated by experimental data generated through extensive simulation from Netaji Subhas Open University, Kolkata, India in terms of its execution times. The proposed FILP models are compared with a commonly used heuristic viz., Integer Linear Programming approach [71] on experimental data which gives an idea about quality of the heuristic. FILP technique is again compared with different AI based heuristic techniques [70] for ETP with respect to best and mean cost as well as execution time measures on Carter benchmark datasets to illustrate its effectiveness. The comparative study is performed using mathematical Model 3 of FILP technique because minimum number of variables is required in its formulation. An appreciable amount of time is required by FILP technique to generate satisfactory solution in comparison to other heuristic solutions. Since ETP is an NP–Hard problem, FILP formulation gives an optimal solution that can serve as good benchmark for other heuristics. The experimental study presented here focuses on producing a methodology that generalizes well over a spectrum of techniques that generates significant results for one or more datasets. The performance of FILP model is finally compared to the best results cited in literature for Carter benchmarks to assess its potential. The problem can be further reduced by formulating with lesser number of allocation variables it without affecting the optimality of the solution obtained. The proposed FLIP model for ETP can also be adapted to solve other ETP as well as several commonly studied combinatorial optimization problems like traveling salesman problem, vehicle routing problem, quadratic assignment problem, large scale job shop problem etc [72–75].

This paper is organized as follows. In next section, examination timetable at Netaji Subhas Open University is discussed involving different hard and soft constraints. This is followed by Mathematical Programming formulation of problem involving various uncertainty measures in section 3. In section 4, simulation results are presented on data obtained from Netaji Subhas Open University. A comparative performance of FILP technique with respect to other AI based heuristic techniques on highly cited Carter datasets is also presented. This is supported by the difference of results of FILP model compared to best cited results. Finally, conclusion and future work are given in section 5.

2. **Examination Timetable at Netaji Subhas Open University.** ETP [2, 4, 18–19] consists in finding exact time allocation within limited time period of number of events (courses to be examined) and assign to them number of resources (invigilators, students and examination rooms) such that different constraints are satisfied. Examination timetable for Netaji Subhas Open University, Kolkata, India is developed keeping in view that no timetable exists for University. The examination period is currently fixed to two weeks with two examination sessions per day. A week is made up of six consecutive days i.e. Monday to Saturday. Some examinations have more candidates than a single room can hold, thus they are scheduled in more than one room. On the other hand, a room can have more than one examination scheduled if sufficient room space is available. To optimize examination space, lecture theatres are also used for examinations. The work is based on these assumptions:(i) An examination is scheduled in any of the available rooms; (ii) All examinations are of variable time length duration; (iii) There are no open book examinations; (iv) The assignment of courses to rooms and timeslots is done as per division of courses in lecture theatres into different groups (v) The maximum mixing of courses in a room is four due to four colors arrangement; (vi) Walking distance between rooms is irrelevant, as there is an interval of at least one hour between examinations. These assumptions closely resemble the actual examination scenario at University.

TABLE 1. Timetable Problem Specifications

| Serial Number | Parameter Description | Quantity |
|---|---|---|
| 1 | Number of Courses | 90 |
| 2 | Number of different Examinations | 200 |
| 3 | Number of scheduled Events | 210 |
| 4 | Number of Semesters | 11 |
| 5 | Type of Examinations (Theory/Laboratory) | 2 |
| 6 | Number of Teachers | 50 |
| 7 | Number of Examination rooms | 19 |
| 8 | Number of Days | 12 |
| 9 | Number of Examination sessions per Day | 2 |

The constraints to be satisfied by timetable are usually divided into two categories viz. *hard* and *soft* constraints [2, 18]. Hard constraints such as avoiding student examination collision and room over-sizing must be satisfied. Soft constraints such as scheduling large candidate examinations [25–26] first may be tolerable but must be minimized as much as possible. Hard constraints are rigidly fulfilled. Such constraints include: (i) No candidate is assigned to more than one examination at same time; (ii) A room cannot be assigned more candidates than its capacity; (iii) No resource are assigned to different events at same time; (iv) Events of same semester are not assigned at same time slot in order for students to write all semester courses; (v) Assigned resources to an event must belong to set of valid resources for that event. In this regard, examination is held in classroom if proper infrastructural arrangements especially for laboratory examinations are there to organize the examination. On the other hand, it is desirable to fulfill soft constraints to the possible extent but is not fully essential for valid solution. Therefore, soft constraints can also be seen as optimization objectives for search algorithm. Such constraints are: (i) Room space wastage is minimized during examinations and each room is utilized as much as possible; (ii) Continuous examinations are discouraged for a candidate in a day; (iii) Each lecturer has at least one time gap between invigilation sessions. The different problem specifications are given in Table 1 and hard and soft constraints considered for this problem are given in Tables 2 and 3 respectively.

TABLE 2. Hard Constraint Specifications

| Serial Number | Hard Constraint |
|---|---|
| 1 | No resource (invigilator, student or examination room) is assigned to different events at same time |
| 2 | Events of same semester are not assigned at same time slot when both events are of type *theory* or when one event is *theory* and other event is *laboratory*. Same semester events held concurrently only if they are both of type *laboratory*, as for each course 4 *laboratory* examinations are scheduled within a week, each appeared by different group of students. |
| 3 | Maximum number of examination sessions per day should not exceed particular value (2) |
| 4 | Each examination session is held in a room belonging to specific set of valid rooms for that examination |
| 5 | Each examination room has its own availability schedule |
| 6 | Each *laboratory* examination is assigned to a teacher that belongs to set of teachers who can conduct the examination |
| 7 | Both *theory* and *laboratory* examinations need two teachers for invigilation purpose |

TABLE 3. Soft Constraint Specifications

| Serial Number | Soft Constraint |
|---|---|
| 1 | Every teacher has his own availability schedule for invigilation ensuring which he submits plan with desirable time periods that suits him best |
| 2 | Every teacher has minimum and maximum limit of invigilation hours which are 12 and 4 respectively during the entire examination period |
| 3 | Travel time of teachers and students between examination rooms within Campus is to be minimized |
| 4 | Room space wastage is minimized during examinations and each room is utilized as much as possible |
| 5 | Continuous examinations are discouraged for a candidate in a day |
| 6 | Each lecturer has at least one time gap between invigilation sessions. |

In Table 1, values 2 for field examination sessions per day denote possible starting examination for each session (from 10:00 am to 5:00 pm) and not complete time slots that can accommodate equal number of consequent examinations. Different examinations have different durations' viz. 2 to 3 hours. Any solution satisfying above constraints is feasible schedule for the problem. The specific case is considered as benchmark and reasons are: (i) The real constraints were easily accessed for developing manual solution to problem in order to set-up examination timetable problem on realistic basis; (ii) There was an easy access to manual solutions for the problem which facilitates making easy comparisons with present results; (iii) The specific problem is generally NP–Hard problem and serves as demanding benchmark for developing an efficient optimization algorithm. There are certain difficulties involved in chosen problem case which are justified by following facts: (i) Problem has two types of examinations viz. *theory* and *laboratory* with diverse characteristics and constraints; (ii) Number of examination rooms is generally small viz. only 19 in University that accommodates all sessions, a fact which makes timetable schedule very tight. Some examination rooms are laboratories designed for *laboratory* examinations and others are *theory* examination rooms. Not all laboratories are occupied by examination sessions for 12 days with only minor time-gaps; (iii) Examinations are held in classrooms as usual except *theory* examinations are assigned to any of the lecture classrooms and *laboratory* examinations are assigned to specific laboratory classrooms; (iv) There are quite large number of teachers each of whom has their own minimum (4) and maximum (12) invigilation sessions during entire examination period. Some teachers are kept as reserved invigilators incase somebody fails to turn up.

3. **Mathematical Programming Formulation of the Problem.** In this section, we discuss mathematical models involving various uncertainty measures in formulating ETP at Netaji Subhas Open University, Kolkata, India keeping in view different assumptions. The uncertainty measures are associated with both hard and soft constraints of problem. Fuzzy Sets [67, 69] are used to model uncertainty and vagueness associated with constraints in final timetable schedule by allowing grades of membership in the set. The model allows decision maker to express his preference to ultimate schedule such that related measure is appropriately represented. Among soft constraints, best availability schedule of each teacher for invigilation, maximum and minimum invigilation workload of each teacher are uncertain due to both human as well as environment factors. In addition, travel time of teachers and students between rooms within campus, time gaps within schedule of each teacher and continuous examinations have an inherent degree of uncertainty and impreciseness factors associated. These constraints are best modeled using Fuzzy Sets [67, 69].

Likewise, hard constraints such as not assignment of events of same semester at similar slot when both events are of type theory or when either event is theory or laboratory, maximum number of examination sessions per day, examination session held in a room belonging to specific set of valid rooms, availability schedule of examination rooms and assignment of laboratory examination to a teacher who can conduct the examination can be modeled through Fuzzy Sets [67, 69]. However, hard constraints like assignment of resources viz. invigilator, student or examination room to different events at same time and requirement of 2 invigilators for theory and laboratory examinations can be effectively modeled through crisp sets. Hard constraints must be satisfied and these are modeled as constraints of problem. Soft constraints are to be minimized and these are modeled as objective function of problem.

**Definition:** A Fuzzy Set $\tilde{A}$ is defined by membership function $\mu_{\tilde{A}}(x)$ which assigns to each object $x$ in universe of discourse $X$ a value representing its grade of membership in Fuzzy Set given by [67, 69],

$$\mu_{\tilde{A}}(x): X \to [0,1] \qquad (1)$$

A variety of shapes are used to represent Fuzzy memberships such as triangular, trapezoidal, bell, s curves etc. Conventionally, choice of curve shape is subjective and allows decision maker to express his preferences.

The estimation of time elapsed with respect to 3$^{rd}$ soft constraint is obtained by taking into consideration nature of teacher, student and location of examination rooms. While some people walk faster, others may walk slowly as a result of which elapsed time are basically dependent on walking speed of different people. Uncertain elapsed times $\tilde{p}_{ij}$ are modeled by using triangular membership functions [67–69] represented by triplet $(p_{ij}^1, p_{ij}^2, p_{ij}^3)$, where $p_{ij}^1$ and $p_{ij}^3$ are lower and upper bounds of elapsed time while $p_{ij}^2$ is modal point as represented in Figure1. The use of triangular fuzzy numbers to model uncertainty in elapsed times may be attributed to the fact that three state representations through triplet $(p_{ij}^1, p_{ij}^2, p_{ij}^3)$ most accurately simulate real life data.

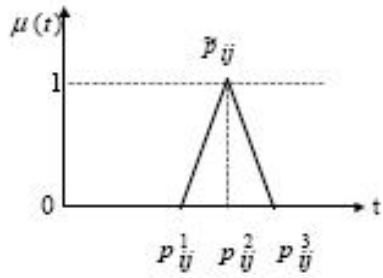

FIGURE 1. Fuzzy representation of Elapsed Times

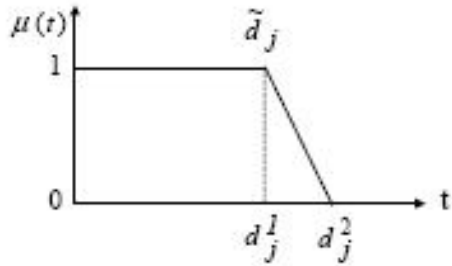

FIGURE 2. Fuzzy representation of Schedule of Teacher

The 2$^{nd}$ soft constraint i.e. weekly invigilation hours of each teacher can similarly be represented by triangular membership functions [67–69] given by triplet $(w_{ij}^1, w_{ij}^2, w_{ij}^3)$, where $w_{ij}^1$ and $w_{ij}^3$ are minimum and maximum bounds of weekly invigilation hours of each teacher while $w_{ij}^2$ is modal point. The 4$^{th}$ soft constraint i.e. room space wastage during examinations and room utilization can effectively be modeled by triplet $(r_{ij}^1, r_{ij}^2, r_{ij}^3)$, where $r_{ij}^1$ and $r_{ij}^3$ are minimum and maximum bounds of room space wastage during examinations and $r_{ij}^2$ is modal point. The 1$^{st}$, 5$^{th}$ and 6$^{th}$ soft constraints are represented using LR trapezoidal membership functions [67–69]. The specific time period must be left between non-contiguous examinations within a day is given by $\tilde{d}_j$ and represented by doublet $(d_j^1, d_j^2)$, where $d_j^1$ and $d_j^2$ denote left and right end of trapezoid as depicted in Figure 2. Likewise, invigilation time gap for each lecturer can similarly be modeled. The invigilation schedule of each teacher with respect to his own availability and desirable time periods that suits him best is given by schedule $\tilde{s}_j$ and represented by doublet $(s_j^1, s_j^2)$, where $s_j^1$ and $s_j^2$ denote left and right end of trapezoid.

The 3$^{rd}$ hard constraint maximum number of examination sessions per day can be represented by triangular membership function [67–69] given by triplet $(e_{ij}^1, e_{ij}^2, e_{ij}^3)$, where $e_{ij}^1$ and $e_{ij}^3$ are minimum and maximum bounds of examination sessions per day while $e_{ij}^2$ is modal point as shown in Figure 1 above. The 5$^{th}$ hard constraint can be represented using LR trapezoidal membership function [67–69]. The availability schedule of each examination room is given by $\tilde{a}_j$ and represented by doublet $(a_j^1, a_j^2)$, where $a_j^1$ and $a_j^2$ denote left and right end of trapezoid as depicted in Figure 2 above. Likewise, 2$^{nd}$, 4$^{th}$ and 6$^{th}$ can similarly be modeled through LR trapezoidal membership function.

The above mentioned hard constraints are observed in existing University examination system. They are strictly considered during problem formulation. Besides these some other hard constraints are also considered while formulating mathematical model of problem. These hard constraints are additional constraints which lead to different mathematical formulations of problem. The major prima face of including these hard constraints is to make the work more practical such that it can be effectively applied to different real world applications. Further reasons and justifications for including these hard constraints are enumerated in following subsections. Based on above discussion different mathematical formulations of problem [71–75] are given in Models 1, 2 and 3.

3.1 **Model 1 for ETP at Netaji Subhas Open University.** In formulating mathematical model 1 for ETP at Netaji Subhas Open University the abovementioned hard and soft constraints are given due consideration. Alongwith this some other hard and soft constraints are considered which reflect real world scenario during examination process at University. These constraints are similarly modeled giving due consideration to different parameters and other aspects. One such parameter is assignment of course to particular time slot in a room. Let this be depicted by $\tilde{x}_{jkr}$ where course $j$ is assigned to time slot $k$ in room $r$. The time slot $k$ takes care of both time and day. As examinations take two weeks spanning over period of twelve days in a semester, they are numbered in increasing order from Monday of first week to Saturday of second week. This gives a total of 24 timeslots for whole examination period. The variable $\tilde{x}_{jkr}$ is Fuzzy number represented through triangular membership function as given in Figure 1 [67–69]. For each feasible solution, there is fuzzy number which is obtained by means of Fuzzy objective function. Hence, to solve this problem for optimal solution as well as Fuzzy value of objective, Fuzzy number ranking method [67–69] obtained from this function is considered which provides different auxiliary conventional optimization model.

3.1.1 **Hard Constraints (Constraints).** The hard constraints of Model 1 are briefly enumerated as follows [2, 18, 72–75]:

(i) Every student is registered for courses that he will take before beginning of semester,

$$\text{i.e., } s_{ij} = \begin{cases} 1, & i^{th} \text{ student} \rightarrow j^{th} \text{ course} \\ 0 & \text{otherwise} \end{cases}$$

(ii) Lecturers are assigned courses from respective departments. It is required to have lecturer identification and courses assigned to each lecturer,

$$\text{i.e., } l_{ij} = \begin{cases} 1, & i^{th} \text{ lecturer} \rightarrow j^{th} \text{ course} \\ 0 & \text{otherwise} \end{cases}$$

(iii) There are $n$ courses scheduled in $m$ rooms and $t$ time slots.

(iv) Examinations are scheduled in rooms whose capacity is known. Let $\tilde{R}_r$ denote examination capacity of room $r$. The examination capacity of room is normally less than teaching capacity so as to disperse candidates. $\tilde{R}_r$ is represented through triangular fuzzy number [67–69] and behaves identically as $\tilde{x}_{jkr}$.

(v) Student is not assigned to more than one examination session at a time. In other words, a student cannot be assigned to two courses $j_1$ and $j_2$ at same time slot $k$,

$$\text{i.e., } s_{ij_1}\tilde{x}_{j_1kr_1} + s_{ij_2}\tilde{x}_{j_2kr_2} \leq 1; k = 1,\ldots,24 \ \forall r_1, \forall r_2, \forall i, \forall j_1, \forall j_2 \text{ such that } j_1 \neq j_2$$

(vi) Examination is not assigned to room which has less capacity than course size; thus courses assigned to time slot $k$ to room $r$ must not exceed examination room capacity, i.e.,

$$\sum_j \sum_i s_{ij}\tilde{x}_{jkr} \leq \tilde{R}_r; k = 1,\ldots,24, r = 1,\ldots,m$$

Thus, if any of the courses $j = 1,2,\ldots,n$ are assigned to time slot $k$ at room $r$, then sum of all students taking courses $j = 1,2,\ldots,n$ must not exceed room capacity. These constraints are considered alongwith hard constraints discussed earlier.

3.1.2 **Soft Constraints (Objective function).** The soft constraints of Model 1 are briefly listed as follows [2, 18, 72–75]:

(i) Room space wastage is minimized during examinations; each room $r$ is utilized as much as possible,

$$\min z = \sum_r \sum_k (\tilde{R}_r - \sum_j \sum_i s_{ij}\tilde{x}_{jkr})$$

(ii) When $i^{th}$ student has consecutive examinations $j_1$ and $j_2$ in time slot $k$ and $k+1$, it is minimized,

$$\min s_{ij_1}\tilde{x}_{j_1kr_1} + s_{ij_2}\tilde{x}_{j_2(k+1)r_2} \ \forall i, \forall k, \forall r_1, \forall r_2, \forall j_1, \forall j_2 \text{ such that } j_1 \neq j_2$$

(iii) When $l^{th}$ lecturer invigilates more than one consecutive examination $j_1$ and $j_2$ in time slots $k$ and $k+1$, it is minimized,

$$\min(l_{ij_1}\tilde{x}_{j_1kr_1} + l_{ij_2}\tilde{x}_{j_2(k+1)r_2}) \ \forall i, \forall j, \forall k, \forall r_1, \forall r_2 \text{ such that } i \neq j$$

(iv) Every teacher has his own availability schedule for invigilation ensuring which he submits plan with desirable time periods that suits him best,

$$\min_{\tilde{s}_j^k} \sum_j \sum_k \tilde{s}_j^k$$

(v) Every teacher has minimum and maximum limit of invigilation hours which are 12 and 4 respectively during entire examination period,

$$\min_{4 \leq \tilde{w}_{ij}^k \leq 12} \sum_i \sum_j \sum_k \tilde{w}_{ij}^k$$

(vi) Travel time of teachers and students between rooms within campus is minimized,

$$\min_{\tilde{p}_{ij}^k} \sum_i \sum_j \sum_k \tilde{p}_{ij}^k$$

Based on above constraints, FILP model [67–69, 72–75] is as follows:

$$\min z = \sum_r \sum_k (\tilde{R}_r - \sum_j \sum_i s_{ij}\tilde{x}_{jkr}) + \sum_i \sum_{j_1} \sum_{j_2} \sum_k \sum_{r_1} \sum_{r_2} (s_{ij_1}\tilde{x}_{j_1kr_1} + s_{ij_2}\tilde{x}_{j_2(k+1)r_2})$$
$$+ \sum_i \sum_{j_1} \sum_{j_2} \sum_k \sum_{r_1} \sum_{r_2} (l_{ij_1}\tilde{x}_{j_1kr_1} + l_{ij_2}\tilde{x}_{j_2(k+1)r_2}) + \min_{\tilde{s}_j^k} \sum_j \sum_k \tilde{s}_j^k$$
$$+ \min_{4 \leq \tilde{w}_{ij}^k \leq 12} \sum_i \sum_j \sum_k \tilde{w}_{ij}^k + \min_{\tilde{p}_{ij}^k} \sum_i \sum_j \sum_k \tilde{p}_{ij}^k$$

subject to $s_{ij_1}\tilde{x}_{j_1kr_1} + s_{ij_2}\tilde{x}_{j_2kr_2} \leq 1; k = 1,\dots\dots,24 \ \forall r_1, \forall r_2, \forall i, \forall j_1, \forall j_2$ such that $j_1 \neq j_2$

$$\sum_j \sum_i s_{ij}\tilde{x}_{jkr} \leq \tilde{R}_r; k = 1,\dots\dots,24, r = 1,\dots\dots,m$$

and other hard constraints

Given timetable with *n* courses, *m* rooms and *t* timeslots, we have a problem with *ntm* variables. A typical examinations timetable at Netaji Subhas Open University involves 250 courses, 50 rooms, 10,000 students and 400 lecturers on 24 slots time interval. This gives a problem with *ntm* = 250 × 24 × 50 = 300000 variables. Thus, problem is generally intractable in nature to be solved by any available software. The number of variables is minimized by breaking variable definition into two separate variables which gives rise to the following mathematical model.

3.2 **Model 2 for ETP at Netaji Subhas Open University.** The mathematical model 2 for ETP at Netaji Subhas Open University is formulated similarly as model 1 given due importance to the abovementioned hard and soft constraints. As discussed previously some other hard and soft constraints are considered which reflect real world examination scenario at University. These constraints are also modeled giving due consideration to different parameters and other aspects. Two such parameters are assignment of course to particular time slot and assignment of course to a room. Let these be depicted by $\tilde{x}_{jk}$ where course *j* assigned to time slot *k* and $\tilde{y}_{jr}$ where course *j* assigned to room *r*. The variables $\tilde{x}_{jk}$ and $\tilde{y}_{jr}$ behave as triangular fuzzy numbers as given in Figure 1 [67–69]. The solution of this problem for optimal solution is obtained through Fuzzy value of objective and Fuzzy number ranking method [67–69] which provides different auxiliary conventional optimization model.

3.2.1 **Hard Constraints (Constraints).** The hard constraints of Model 2 are briefly enumerated as follows [2, 18, 72–75]:

(i) The $i^{th}$ student cannot have more than one course scheduled at same time slot $k$,

i.e., $s_{ij_1}\tilde{x}_{j_1k} + s_{ij_2}\tilde{x}_{j_2k} \leq 1; k = 1,..............,24 \ \forall i, \forall j_1, \forall j_2$ such that $j_1 < j_2$

(ii) Room $r$ cannot have more students than its capacity $\tilde{R}_r$,

i.e., $\sum_j \sum_i s_{ij}\tilde{y}_{jr} \leq \tilde{R}_r; r = 1,................,m$

These constraints are considered alongwith hard constraints discussed earlier.

3.2.2 **Soft Constraints (Objective function).** The soft constraints of Model 2 are briefly listed as follows [2, 18, 72–75]:

(i) Lecturer $i$ cannot invigilate consecutive examination $j_1$ and $j_2$; there is at least one gap between $[k, k+1]$ of an examination slot,

$\min l_{ij_1}\tilde{x}_{j_1k} + l_{ij_2}\tilde{x}_{j_2(k+1)} \ \forall i, \forall k, \forall j_1, \forall j_2$ such that $j_1 < j_2$

(ii) When student $i$ has two or more consecutive examination $j_1$ and $j_2$, it is minimized,

$\min s_{ij_1}\tilde{x}_{j_1k} + s_{ij_2}\tilde{x}_{j_2(k+1)} \ \forall i, \forall k, \forall j_1, \forall j_2$ such that $j_1 < j_2$

(iii) Room space wastage is minimized, $\tilde{R}_r - \sum_j \sum_i (s_{ij}\tilde{y}_{jr}); r = 1,............,m$

(iv) Every teacher has his own availability schedule for invigilation ensuring which he submits plan with desirable time periods that suits him best,

$\min_{\tilde{s}_j^k} \sum_j \sum_k \tilde{s}_j^k$

(v) Every teacher has minimum and maximum limit of invigilation hours which are 12 and 4 respectively during entire examination period,

$\min_{4 \leq \tilde{w}_{ij}^k \leq 12} \sum_i \sum_j \sum_k \tilde{w}_{ij}^k$

(vi) Travel time of teachers and students between rooms within campus is minimized,

$\min_{\tilde{p}_{ij}^k} \sum_i \sum_j \sum_k \tilde{p}_{ij}^k$

Considering the above constraints, FILP model [67–69, 72–75] is as follows:

$\min z = \sum_r (\tilde{R}_r - \sum_j \sum_i s_{ij}\tilde{y}_{jr}) + \sum_i \sum_{j_1} \sum_{j_2} \sum_k (s_{ij_1}\tilde{x}_{j_1k} + s_{ij_2}\tilde{x}_{j_2(k+1)})$

$+ \sum_i \sum_{j_1} \sum_{j_2} \sum_k (l_{ij_1}\tilde{x}_{j_1k} + l_{ij_2}\tilde{x}_{j_2(k+1)}) + \min_{\tilde{s}_j^k} \sum_j \sum_k \tilde{s}_j^k$

$+ \min_{4 \leq \tilde{w}_{ij}^k \leq 12} \sum_i \sum_j \sum_k \tilde{w}_{ij}^k + \min_{\tilde{p}_{ij}^k} \sum_i \sum_j \sum_k \tilde{p}_{ij}^k$

subject to $s_{ij_1}\tilde{x}_{j_1k} + s_{ij_2}\tilde{x}_{j_2k} \leq 1; k = 1,..............,24 \ \forall i, \forall j_1, \forall j_2$ such that $j_1 < j_2$

$$\sum_j \sum_i s_{ij} \tilde{y}_{jr} \leq \tilde{R}_r; r = 1,\ldots\ldots\ldots,m$$

and other hard constraints

In this model, we have $n(t+m) = 250 \times (24 + 50) = 18{,}500$ variables. This is a considerable reduction from previous model. However, number of variables is still large enough to solve real timetable problem. The number of variables can be further reduced if the number of crisp as well as the fuzzy variables is minimized.

**3.3 Model 3 for ETP at Netaji Subhas Open University.** Finally, the mathematical model 3 for ETP at Netaji Subhas Open University is formulated similarly as models 1 given due importance to the abovementioned hard and soft constraints. As mentioned previously some other hard and soft constraints are considered which reflect real world examination scenario at University. These constraints are also modeled giving due consideration to different parameters and other aspects. Three such parameters are student taking a course, assignment of lecture to a course and capacity of room. Let these be depicted by $\tilde{G}_{si}$ where student $s$ takes course $i$, $\tilde{A}_{li}$ where lecture $l$ assigned to course $i$ and $\tilde{R}_i$ which denotes capacity of room $i$. The variables $\tilde{G}_{si}$, $\tilde{A}_{li}$ and $\tilde{R}_i$ behave as triangular fuzzy numbers as given in Figure 1 [67–69]. Other important parameters are $T_c$ which denote the time slot in which course $c$ is slotted and $C_i$ which denote room assigned for course $i$. The solution of this problem for optimal solution is obtained through Fuzzy value of objective and Fuzzy number ranking method [67–69] which provides a different auxiliary conventional optimization model.

**3.3.1 Hard Constraints (Constraints).** The hard constraints of Model 3 are briefly enumerated as follows [2, 18, 72–75]:

(i) Courses $i$ and $j$ registered by same student $s$ cannot be scheduled in same slot as it may result in student collision,

$$T_i \tilde{G}_{si} \neq T_j \tilde{G}_{sj}, \forall s, \forall i, j \ni i \prec j$$

(ii) Count of all students $s$ registered for courses $i$ which have been slotted in room $r$ must not exceed capacity of room,

$$\sum_{i \ni C_i = r} \sum_s \tilde{G}_{si} \leq \tilde{R}_r, \forall r$$

Since, $C_i$ is variable, it is to be removed from bounds which are done by introducing 0/1 value $\delta_{ir}$, such that

$$\delta_{ir} = \begin{cases} 1, C_i = r \\ 0, otherwise \end{cases}$$

Thus, the above expression can be rewritten as $\sum_i \sum_s \delta_{ir} \tilde{G}_{si} \leq \tilde{R}_r, \forall r$.

(iii) To enforce 0/1 variable constraint $C_i - \delta_{ir} \prec r, \forall i, \forall r$ is introduced.

(iv) Given a course slotted in room $r$ i.e. $C_r$ where room $r$ has no standby generator, then time in which this course is slotted cannot be multiple of 3 i.e. it cannot be held in evening hours i.e., $T_{C_r} \neq p, p \in E, r \notin G$. These constraints are considered alongwith hard constraints discussed earlier.

3.3.2 **Soft Constraints (Objective function).** The soft constraints of Model 3 are briefly listed as follows [2, 18, 72–75]:

(i) Room space wastage is minimized, i.e., $\sum_r (\tilde{R}_r - \sum_{i \ni C_i = r} \sum_s \tilde{G}_{si})$.

(ii) Consecutive examinations for a student is avoided by considering gap between two examinations at least 2,
$$|T_i \tilde{G}_{si} - T_j \tilde{G}_{sj}| \succ 2, \forall s, \forall i, j \ni i \prec j$$
The above expression can also be rewritten as,
$$\min(2 - (T_i \tilde{G}_{si} - T_j \tilde{G}_{sj})), \forall s, \forall i, j \ni i \prec j$$
$$\text{or } \min \sum_s \sum_{i \prec j} (2 - (T_i \tilde{G}_{si} - T_j \tilde{G}_{sj}))$$

(iii) Every teacher has his own availability schedule for invigilation ensuring which he submits plan with desirable time periods that suits him best,
$$\min_{\tilde{s}_j^k} \sum_j \sum_k \tilde{s}_j^k$$

(iv) Every teacher has minimum and maximum limit of invigilation hours which are 12 and 4 respectively during entire examination period,
$$\min_{4 \leq \tilde{w}_{ij}^k \leq 12} \sum_i \sum_j \sum_k \tilde{w}_{ij}^k$$

(v) Travel time of teachers and students between rooms within campus is minimized,
$$\min_{\tilde{p}_{ij}^k} \sum_i \sum_j \sum_k \tilde{p}_{ij}^k$$

Keeping in view the above constraints, FILP model [67–69, 72–75] is as follows:
$$\min z = \sum_r (\tilde{R}_r - \sum_i \sum_s \delta_{ir} G_{si}) + \sum_s \sum_{i \prec j} (2 - (T_i \tilde{G}_{si} - T_j \tilde{G}_{sj})) + \min_{\tilde{s}_j^k} \sum_j \sum_k \tilde{s}_j^k$$
$$+ \min_{4 \leq \tilde{w}_{ij}^k \leq 12} \sum_i \sum_j \sum_k \tilde{w}_{ij}^k + \min_{\tilde{p}_{ij}^k} \sum_i \sum_j \sum_k \tilde{p}_{ij}^k$$
$$\text{subject to } T_i \tilde{G}_{si} - T_j \tilde{G}_{sj} \neq 0, \forall s, \forall i, j \ni i \prec j$$
$$\sum_i \sum_s \delta_{ir} \tilde{G}_{si} \leq \tilde{R}_r, \forall r$$
$$C_i - \delta_{ir} \prec r, \forall i, \forall r$$
$$T_{C_r} \neq p, p \in E, r \notin G$$
$$\delta \in \{0,1\}, T, C \in I$$
and other hard constraints

In this model, there are three types of variables viz., $T, C$ and $\delta$ giving total of $t + n + nm = 28 + 200 + (200 \times 50) = 10,228$. This is significant reduction in number of variables than previous two models. However, number of variables is still large to be solved. This is obvious since problem is NP–Hard. Small instances of problems can be solved to optimality which can act as benchmarks for testing performance of heuristics.

4. **Simulation Results and Discussions.** This section discusses performance of proposed different FILP techniques for ETP on experimental data obtained from Netaji Subhas Open University, Kolkata, India. This data is generated through extensive simulation of the existing system at University. The effectiveness of the mathematical models of FILP technique in section 3 are further demonstrated by comparing with different AI based heuristic techniques on 13 Carter benchmark datasets. Section 4.1 discusses results obtained on data from Netaji Subhas Open University, Kolkata, India. A comparative performance of FILP technique with respect to other AI based heuristic techniques on highly cited Carter datasets is presented in section 4.2. Alongwith this difference of results of FILP technique as compared to best cited results is also illustrated.

4.1 **Performance of FILP technique on test data from Netaji Subhas Open University.** The proposed FILP technique is implemented in Microsoft Visual C++ 6 under Windows XP on 2 GHz Intel Core 2 Duo processor having 512 RAM and 1 GB of memory. Test data are considered from departments of Mathematics, Computer Science, Physics, Life Science, Management, English and Bengali. The results for mathematical models 1, 2 and 3 discussed in section 3 are summarized in tables 4, 5 and 6 respectively.

A small data set is used as number of constraints grows exponentially which increases the size of problem. The solution is found in Model 2 after lesser number of iterations than Model 1. The number of variables and constraints is significantly lower. However, solution times are not directly proportional to size of problem. Management data in Model 2 took longer than Computer Science data on same model despite of the fact that two departments have same number of variables. This is generally due to the fact that performance of algorithm also depends on other factors such as deeper constraints are towards optimal polytope. Using polyhedral combinatorics [30–31, 72, 74] it is worthwhile to find facets or deepest cuts associated with such model. This will trim down infeasibilities and make it easy for branch and bound procedure to solve the remaining problem. Such an approach brings down the size of problem that could be solved by Mathematical Programming [72–75]. This has been applied with success to several Linear Programming models [72–75]. Generally two models give same solution but Model 2 is smaller in size and could be used to solve bigger problems as it gives set of optimal solutions which can test the performance of heuristics. Further on comparing result of Models 1 and 2 it is found that execution time of Model 2 is lower than that of Model 1. This difference generally exists due to the fact that number of variables and constraints in Model 2 are less than those of Model 1. Finally in Model 3 solution is found after lesser number of iterations than Models 1 and 2. Also number of variables and constraints is reduced significantly [72–75]. While comparing Model 3 with Models 1 and 2 it is found that infeasibilities are cut down making it better candidate for branch and bound procedure than other models. Generally three models give same solution but

Model 3 is smaller in size gives set of optimal solutions and can be used to solve bigger real life problem sizes which can be effectively used to test performance of heuristics.

TABLE 4. Test results for FILP Model 1

| FILP Model 1 | | | |
|---|---|---|---|
| Departments | Variables | Constraints | Execution Time (seconds) |
| Mathematics | 336 | 72,666 | 24.46 |
| Computer Science | 446 | 112,025 | 26.36 |
| Physics | 336 | 86,769 | 25.19 |
| Life Science | 336 | 164,396 | 27.46 |
| Management | 446 | 110,019 | 26.30 |
| English | 286 | 66,889 | 22.18 |
| Bengali | 286 | 64,886 | 22.18 |

TABLE 5. Test results for FILP Model 2

| FILP Model 2 | | | |
|---|---|---|---|
| Departments | Variables | Constraints | Execution Time (seconds) |
| Mathematics | 124 | 7,786 | 24.16 |
| Computer Science | 127 | 6,669 | 24.07 |
| Physics | 124 | 8,968 | 24.25 |
| Life Science | 124 | 9,846 | 24.44 |
| Management | 127 | 10,896 | 25.45 |
| English | 107 | 5,787 | 21.07 |
| Bengali | 107 | 4,996 | 21.69 |

TABLE 6. Test results for FILP Model 3

| FILP Model 3 | | | |
|---|---|---|---|
| Departments | Variables | Constraints | Execution Time (seconds) |
| Mathematics | 48 | 255 | 12.37 |
| Computer Science | 44 | 290 | 9.96 |
| Physics | 44 | 346 | 11.33 |
| Life Science | 48 | 396 | 12.86 |
| Management | 44 | 425 | 12.46 |
| English | 27 | 227 | 7.66 |
| Bengali | 27 | 207 | 7.37 |

A comparative study of Models 1, 2 and 3 with respect to heuristic Integer Linear Programming model [72–75] are given in Tables 7, 8 and 9. On comparing results in Tables 4, 5 and 6 with Tables 7, 8 and 9 it is evident that execution times are lower than for Models 1, 2 and 3 with respect to heuristics. This is mainly because various allocation variables associated with Models 1, 2 and 3 are formulated using fuzzy numbers [67–69] which handles vagueness and imprecision involved in problem as a result of which feasible solution is obtained in less time. The rigidity involved in allocation variables of Integer Linear Programming heuristic [72–75] mainly attributes higher execution times.

TABLE 7. Test results for Model 1 versus Integer Linear Programming Heuristic

| Model 1 versus Integer Linear Programming Heuristic | | | |
|---|---|---|---|
| Departments | Variables | Constraints | Execution Time (seconds) |
| Mathematics | 336 | 72,666 | 26.55 |
| Computer Science | 446 | 112,025 | 28.90 |
| Physics | 336 | 86,769 | 27.69 |
| Life Science | 336 | 164,396 | 29.98 |
| Management | 446 | 110,019 | 28.56 |
| English | 286 | 66,889 | 24.75 |
| Bengali | 286 | 64,886 | 24.79 |

TABLE 8. Test results for Model 2 versus Integer Linear Programming Heuristic

| Model 2 versus Integer Linear Programming Heuristic | | | |
|---|---|---|---|
| Departments | Variables | Constraints | Execution Time (seconds) |
| Mathematics | 124 | 7,786 | 26.36 |
| Computer Science | 127 | 6,669 | 26.09 |
| Physics | 124 | 8,968 | 26.37 |
| Life Science | 124 | 9,846 | 26.86 |
| Management | 127 | 10,896 | 32.64 |
| English | 107 | 5,787 | 26.05 |
| Bengali | 107 | 4,996 | 24.69 |

TABLE 9. Test results for Model 3 versus Integer Linear Programming Heuristic

| Model 3 versus Integer Linear Programming Heuristic | | | |
|---|---|---|---|
| Departments | Variables | Constraints | Execution Time (seconds) |
| Mathematics | 48 | 255 | 14.75 |
| Computer Science | 44 | 290 | 11.96 |
| Physics | 44 | 346 | 13.37 |
| Life Science | 48 | 396 | 14.98 |
| Management | 44 | 425 | 14.69 |
| English | 27 | 227 | 10.75 |
| Bengali | 27 | 207 | 10.37 |

4.2 **Comparison of FILP technique with other Techniques.** The proposed FILP technique is tested on data instances prepared by Carter, Laporte and Lee in 1996 on 13 real world examination timetable problems from 3 Canadian High Schools, 5 Canadian, 1 American, 1 British and 1 Middle-East Universities which is readily available on website http://www.cs.nott.ac.uk/~rxq/data.htm. The datasets measure performance of approaches related to ETP and is prepared carefully to mimic real word ETP at Netaji Subhas Open University, Kolkata with different size and supersets of constraints. The datasets are of varying sizes with respect to different parameter values. Since FILP is basically a Fuzzy based optimization technique it is worthwhile to compare it with other AI based heuristic techniques for ETP to illustrate its effectiveness. To demonstrate significance of FILP technique, a comparative performance of best and mean cost as well as execution times is made with respect to seven different AI based heuristics when applied to Carter benchmark datasets. These results are given in Tables 10 and 11 respectively.

TABLE 10. Comparison of best and mean cost of Fuzzy Integer Linear Programming (Model 3) technique with other AI based Heuristic techniques

| Datasets | M1 | M2 | M3 | M4 | M5 | M6 | M7 | FILP-M3 |
|---|---|---|---|---|---|---|---|---|
| CAR–F–92 | 6.21 | 4.28 | 4.80 | 4.40 | 4.54 | 4.38 | 4.31 | 4.27 |
|  | 6.25 | 4.36 | 4.86 | 4.42 | 4.60 | 4.40 | 4.32 | 4.31 |
| CAR–S–91 | 7.01 | 4.97 | 5.70 | 5.20 | 5.29 | 5.08 | 5.24 | 4.95 |
|  | 7.07 | 5.10 | 5.82 | 5.30 | 5.37 | 5.18 | 5.30 | 4.96 |
| EAR–F–83 | 42.81 | 36.86 | 36.86 | 36.90 | 37.02 | 38.44 | 37.75 | 36.78 |
|  | 44.24 | 37.22 | 36.96 | 34.98 | 37.18 | 38.45 | 37.79 | 36.80 |
| HEC–S–92 | 12.90 | 11.85 | 11.90 | 12.30 | 12.79 | 11.96 | 11.86 | 11.85 |
|  | 12.96 | 11.85 | 11.97 | 12.31 | 12.80 | 11.97 | 11.87 | 11.86 |
| KFU–S–93 | 18.47 | 14.62 | 15.00 | 14.64 | 15.81 | 14.67 | 14.96 | 14.50 |
|  | 18.50 | 14.62 | 15.04 | 14.67 | 15.84 | 14.70 | 14.99 | 14.52 |
| LSE–F–91 | 15.62 | 11.14 | 12.10 | 11.20 | 13.46 | 11.69 | 12.80 | 11.14 |
|  | 15.64 | 11.14 | 12.12 | 11.21 | 13.47 | 11.70 | 12.86 | 11.17 |
| PUR–S–93 | 7.92 | 4.73 | 5.40 | - | - | - | 4.90 | 4.72 |
|  | 7.99 | 4.78 | 5.44 | - | - | - | 4.97 | 4.75 |
| RYE–S–93 | 10.50 | 9.65 | 10.20 | 9.70 | 10.39 | 9.89 | 9.72 | 9.65 |
|  | 10.54 | 9.70 | 10.26 | 9.75 | 10.44 | 9.90 | 9.79 | 9.66 |
| STA–F–83 | 161.00 | 158.33 | 158.40 | 159.20 | 160.75 | 158.72 | 158.37 | 158.30 |
|  | 161.07 | 158.33 | 158.47 | 159.27 | 160.76 | 158.75 | 158.45 | 158.31 |
| TRE–S–92 | 10.98 | 8.48 | 8.80 | 8.48 | 9.31 | 8.78 | 8.89 | 8.37 |
|  | 11.04 | 8.52 | 8.84 | 8.48 | 9.37 | 8.87 | 8.90 | 8.37 |
| UTA–S–92 | 4.76 | 3.40 | 3.80 | 3.60 | 3.57 | 3.55 | 3.44 | 3.35 |
|  | 4.80 | 3.43 | 3.84 | 3.64 | 3.60 | 3.62 | 3.46 | 3.37 |
| UTE–S–92 | 29.69 | 28.88 | 28.97 | 29.00 | 30.32 | 29.63 | 29.07 | 28.86 |
|  | 29.70 | 28.88 | 28.98 | 29.02 | 30.32 | 29.65 | 29.10 | 28.87 |
| YOR–F–83 | 40.83 | 40.74 | 41.60 | 41.20 | 40.80 | 40.85 | 40.77 | 40.72 |
|  | 40.84 | 41.52 | 41.96 | 41.28 | 40.87 | 40.86 | 40.80 | 40.72 |

In Tables 7 and 8, the experiments are performed on different datasets of varying sizes. In this process, the following different AI based heuristic abbreviations are used: M1: Roulette Wheel Graph Coloring Technique [34]; M2: Heuristic Combinations for Hyper Heuristic Technique [50]; M3: Ant Colonization Technique [47]; M4: Ahuja–Orlin Technique [48]; M5: Fuzzy Logic Technique [51]; M6: Ordering Heuristics Technique [52]; M7: Decision Tree Based Routine Generation Technique [53]; FILP-M3: Fuzzy Integer Linear Programming Technique (Model 3). The mathematical model 3 of FILP is used because minimum number of variables is required in its formulation. The cost for each timetable is calculated using proximity cost function defined by the following equation [50]. It assesses the quality of a timetable in terms of how well examinations are spread. The cost of this function is minimized for each ETP.

$$\frac{\sum_{i,j} w(|e_i - e_j|) N_{ij}}{S}$$

where, $|e_i - e_j|$ is distance between the periods for each pair of examinations $(e_i, e_j)$ with common students; $N_{ij}$ is number of students common to both examinations; $S$ is total number of students and $w(1) = 16$, $w(2) = 8$, $w(3) = 4$, $w(4) = 2$, $w(5) = 1$, i.e. smaller the distance between periods the higher the weight allocated. When $n > 5$ $w(n) = 0$. The best solutions generated are accessible from http://saturn.cs.unp.ac.za/~nelishiap/et/heuristics.htm. The proximity costs of each of these solutions are highlighted in Table 10. The mean cost for problem is the average of proximity cost of the best solution obtained for each of forty runs performed. The overall system did not include mechanisms to ensure that feasible timetables were produced. However, different heuristic techniques produced feasible timetables for all benchmarks.

TABLE 11. Comparison of Execution Times (in minutes) of Fuzzy Integer Linear Programming (Model 3) technique with other AI based Heuristic techniques

| Datasets | M1 | M2 | M3 | M4 | M5 | M6 | M7 | FILP-M3 |
|---|---|---|---|---|---|---|---|---|
| CAR–F–92 | 10 | 7 | 9 | 8 | 9.50 | 8 | 7 | 7 |
| CAR–S–91 | 12 | 10 | 11 | 10 | 10 | 10 | 10 | 10 |
| EAR–F–83 | 4 | 2 | 2 | 1.50 | 2 | 3 | 2 | 1.50 |
| HEC–S–92 | 14 | 11 | 10 | 9.50 | 11 | 11 | 11 | 9.55 |
| KFU–S–93 | 12 | 5 | 6 | 5.50 | 7 | 5 | 6 | 5 |
| LSE–F–91 | 5 | 2 | 3 | 2 | 4 | 3 | 3.50 | 2 |
| PUR–S–93 | 110 | 80 | 90 | - | - | - | 87 | 80 |
| RYE–S–93 | 5 | 5 | 5 | 5 | 5.50 | 5 | 5 | 5 |
| STA–F–83 | 31 | 0.50 | 0.70 | 0.80 | 0.90 | 0.55 | 0.50 | 0.50 |
| TRE–S–92 | 3 | 2 | 2 | 2 | 2.50 | 2 | 2 | 2 |
| UTA–S–92 | 10 | 9 | 9.50 | 9 | 9 | 9 | 9 | 9 |
| UTE–S–92 | 2 | 1 | 1 | 1 | 1.20 | 0.90 | 1 | 1 |
| YOR–F–83 | 3 | 2 | 2 | 2 | 2 | 3 | 2 | 2 |

From Tables 10 and 11, it is evident that different AI based techniques performed appreciable results on different data sets. Experiments are performed on four small datasets, four medium datasets and five large datasets. They are tested using 40 runs for each instance. The proposed FILP-M3 technique produced the best overall results performing just as good as or better than other heuristic techniques for 13 Carter benchmarks. Time consumed for each dataset is variable in nature and depends upon different parameters involved in each technique. The execution time for each heuristic is more or less the same for all techniques. FILP technique takes smaller amount of time to generate satisfactory solution in comparison to other heuristic solutions. Although the study presented in this work focuses on developing methodology that generalizes well over spectrum of techniques that produces significant results for one or more datasets, the performance of this method is compared to the best results cited in literature for Carter benchmarks to assess potential of this methodology. This fact is illustrated by tabulating difference from best cited results as given in Table 12. It is evident that even though the method described in this paper only performs construction phase and not an improvement phase, results produced by FILP-M3 technique is comparable to the best results cited in literature for Carter benchmarks. Furthermore, FILP-M3 method has produced better results than some of methodologies and outperformed on number of benchmarks. The results presented in this section show effectiveness and potential of FILP-M3 technique as general methodology for producing good quality solutions to ETP at Netaji Subhas Open University, Kolkata, India.

TABLE 12. Comparison of the results obtained by FILP (Model 3) technique and best result cited in literature

| Datasets | Fuzzy Integer Linear Programming (Model 3) | Best result cited | Difference |
|---|---|---|---|
| CAR–F–92 | 4.27 | 4.28 | 0.01 |
| CAR–S–91 | 4.95 | 4.97 | 0.02 |
| EAR–F–83 | 36.78 | 36.86 | 0.08 |
| HEC–S–92 | 11.85 | 11.85 | 0.00 |
| KFU–S–93 | 14.50 | 14.62 | 0.12 |
| LSE–F–91 | 11.14 | 11.14 | 0.00 |
| PUR–S–93 | 4.72 | 4.73 | 0.01 |
| RYE–S–93 | 9.65 | 9.65 | 0.00 |
| STA–F–83 | 158.30 | 158.33 | 0.03 |
| TRE–S–92 | 8.37 | 8.48 | 0.11 |
| UTA–S–92 | 3.35 | 3.40 | 0.05 |
| UTE–S–92 | 28.86 | 28.88 | 0.02 |
| YOR–F–83 | 40.72 | 40.74 | 0.02 |

Since FILP approach to ETP is basically a heuristic solution, it can effectively be applied to other optimization problems like lot sizing and pricing multi-product, multi-period with discrete time model [15], vehicle routing problem [14], multi-objective optimization problems [13] and scheduling large-scale job shops problem [16]. The problem parameters can effectively be modeled through proposed FILP models using different Fuzzy membership functions. This can improve the optimality of solution of these problems by taking care of inherent uncertainty and vagueness involved. In simulating optimization algorithm [14], matrix analysis and their transformations can easily be modeled through either of the FILP models and then applied to solve vehicle routing problem. They can also be integrated with other techniques such as Genetic Algorithms to improve quality of solution. This hybrid technique can also be applied to other combinatorial optimization problems like traveling salesman problem, quadratic assignment problem etc to produce optimal solutions. Likewise, the Evolutionary Algorithm for multi-objective optimization problem [13] can be hybridized with FILP models to produce promising results. The constraints in multi-objective optimization problem can be formulated using FILP models. The fitness function can also be optimized through Fuzzy membership functions. Finally, FLIP models can effectively applied to scheduling large scale job shops such that total weighted tardiness is minimized. The sub-problems defined through Simulated Annealing approach and then solved using hybridized FILP based Particle Swarm Optimization Algorithm. This will improve optimization efficiency, jobs' bottleneck characteristic values which are utilized as an immune mechanism to guide sub-problem solving process. By integrating FILP with abovementioned techniques quality of solution and execution times are also reduced.

5. **Conclusion and Future Work.** In this work three FILP mathematical models are presented for ETP. Many possible solutions exist for ETP in literature. However, there is an inherent degree of impreciseness and vagueness involved in both hard and soft constraints of the problem. These uncertainties are effectively handled using Fuzzy Sets by allowing grades of membership in the set underlying different assumptions. The model allows decision maker to express his preference to ultimate schedule such that related measure is appropriately represented. The solution to problem is obtained using Fuzzy number ranking method. Each feasible solution has Fuzzy number obtained by Fuzzy objective function. The performance of different FILP techniques are demonstrated by experimental data generated through extensive simulation from Netaji Subhas Open University, Kolkata, India in terms of its execution times. The proposed FILP models are also compared with a commonly used heuristic viz., Integer Linear Programming approach on experimental data which gives an idea about quality of heuristic. FILP is basically a Fuzzy based optimization technique hence it is again compared with different AI based heuristic techniques for ETP with respect to best and mean cost as well as execution time measures on Carter benchmark datasets to illustrate its effectiveness. The comparative study is performed using mathematical Model 3 of FILP technique because minimum number of variables is required in its formulation. FILP technique takes an appreciable amount of time to generate satisfactory solution in comparison to other heuristic solutions. Since ETP is NP–Hard problem, FILP formulation gives an optimal solution that can serve as good benchmark for other heuristics. The experimental study presented here focuses on producing a methodology that generalizes well over a spectrum of techniques that generates significant results for one or more datasets. The performance of FILP model is finally compared to the best results cited in literature for Carter benchmarks to assess its potential. The problem can be further reduced by formulating with lesser number of allocation variables it without affecting the optimality of solution obtained. Future investigation will encompass automating examination timetable of University through hybrid methodologies by integrating with other AI based paradigms such as Artificial Neural Networks, Genetic Algorithms, Rough Sets and Evolutionary Algorithms. This will also allow for heuristic combinations to be tailored to each of the different problem domains. The proposed FLIP model for ETP can also be adapted to solve other ETP as well as combinatorial optimization problems like traveling salesman problem, vehicle routing problem, quadratic assignment problem, large scale job shop problem as well as other multi-objective optimization problems.